\documentclass[11pt,a4paper]{article}
\usepackage{times,latexsym}
\usepackage{url}
\usepackage[T1]{fontenc}

\usepackage[]{tacl2018v2}

\usepackage{xspace,mfirstuc,tabulary}

\newif\iftaclinstructions
\taclinstructionsfalse 
\iftaclinstructions
\renewcommand{\confidential}{}
\renewcommand{\anonsubtext}{(No author info supplied here, for consistency with
TACL-submission anonymization requirements)}
\newcommand{\instr}
\fi

\iftaclpubformat 

\else

\fi

\usepackage{pbox}
\usepackage{times}
\usepackage{amsfonts}
\usepackage[normalem]{ulem}
\usepackage{amssymb}
\usepackage{textcomp}
\usepackage{graphicx}
\usepackage{amsmath}
\usepackage{csquotes}
\usepackage{multirow}
\usepackage{url}
\usepackage{bm}
\usepackage{booktabs}
\usepackage{color}
\usepackage{tikz}

\newcommand{\rdep}[1]{\ $\xrightarrow{\text{\tiny #1}}$\ }

\title{Human acceptability judgements for extractive sentence compression}

\taclpubformattrue 

\author{
  Abram Handler$^\dagger$, Brian Dillon$^\ast$ and Brendan O'Connor$^\dagger$ \\
   College of Information and Computer Sciences$^\dagger$ \\
   Department of Linguistics$^\ast$ \\
   University of Massachusetts Amherst \\
    { \texttt{ahandler@cs.umass.edu}} \\
}

\date{}

\begin{document}
\maketitle
\begin{abstract}
Recent approaches to English-language sentence compression rely on parallel corpora consisting of sentence--compression pairs. However, a sentence may be shortened in many different ways, which each might be suited to the needs of a particular application. Therefore, in this work, we collect and model crowdsourced judgements of the acceptability of many possible sentence shortenings. We then show how a model of such judgements can be used to support a flexible approach to the compression task. We release our model and dataset for future work.
\end{abstract}


\section{Introduction}\label{s:intro}

In natural language processing, \textit{sentence compression} refers to the task of automatically shortening a longer sentence \cite{Knight2000StatisticsBasedS,clarke2008global,filippova2015sentence}. Traditional approaches attempt to create compressions which $(i)$ maintain readability, $(ii)$ retain the most important information from the source sentence and $(iii)$ achieve some fixed or flexible rate of compression \cite{napoles2011evaluating}.

However, in practice, applications which make use of sentence compression techniques will require dramatically different strategies on how to best achieve these three competing goals: a journalist might want abbreviated quotes from politicians, while a bus commuter might want movie review snippets on their mobile phone. Each such application presents $(i)$ different readability requirements, $(ii)$ different definitions of importance and $(iii)$ different brevity constraints. 

Thus, while recent research into extractive sentence compression often assumes a single, best shortening of a sentence \cite{napoles2011evaluating}, in this work we argue that any compression which matches the readability, informativeness and brevity requirements of a given application is a plausible shortening.
Practical compression methods will need to identify the best possible shortening for a given application, not just recover the ``gold standard'' compression.

\begin{table*}[ht!]
\begin{center}
\begin{tabular}{p{14cm}}
\textbf{Sentence.} Pakistan launched a search for its missing ambassador to Afghanistan on Tuesday, a day after he disappeared in a Taliban area. \\
\textbf{Headline.} Pakistan searches for missing ambassador.  \\
\textbf{``Gold'' compression.} Pakistan launched a search for its missing ambassador. \\
\textbf{Alternate 1.} Pakistan launched a search for its missing ambassador to Afghanistan on Tuesday. {\small($\textsc{A}(c)$ = -1.367, \textsc{Brevity} = 84 characters max., \textsc{Importance} = 1)} \\
\textbf{Alternate 2.} Pakistan launched search Tuesday. {\small ($\textsc{A}(c)$ = -6.144, \textsc{Brevity} = 59 characters max., \textsc{Importance} = 0)} \\
\end{tabular}
\end{center}
\caption{A sentence, headline and ``gold compression'' from a standard sentence compression dataset \cite{filippova2013overcoming}, along with two alternate compressions constructed with a system supervised with human acceptability judgements (\S\ref{s:extrinsic}). The alternate compressions reflect different  \textsc{Brevity} requirements and different adherence to an application-specific \textsc{Importance} criterion. In this case, the brevity requirement is expressed with a hard maximum character constraint and the importance criterion is expressed with a binary score, indicating if a sentence includes a query term ``Afghanistan''. We use our $\textsc{A}(c) \in (-\infty, 0]$ metric (\S\ref{s:extrinsic}) to measure the \textsc{Acceptability} of each compression; a higher score indicates a compression is more likely to be well-formed. Alternate 2 is neither entirely garbled nor perfectly well-formed, reflecting the gradient-based nature of acceptability (\S\ref{s:acceptability}).}
\label{t:all_compressions}
\end{table*}


Traditional supervision for the compression task does not offer an obvious method for achieving this application-specific objective. Standard supervision consists of individual sentences paired with ``gold standard'' compressions \cite{filippova2013overcoming}: offering just one reasonable shortening for each sentence in the corpus, from among the many plausible compressions.
Table \ref{t:all_compressions} demonstrates this point in detail.
Therefore, in this work, we: 

\begin{itemize}
\item{Collect and release\footnote{\url{http://slanglab.cs.umass.edu/compression}} a large, crowdsourced dataset of human acceptability judgements \cite{sprouse2011validation}, specifically tailored for the sentence compression task (\S\ref{s:acceptability}). 
Acceptability judgments are native speakers' self-reported perceptions of the well-formedness of a sentence \cite{shutzesprouse}.}
\item{Present and evaluate a model of these acceptability judgements (\S\ref{s:intrinsic}).}
\item{Use this model to define an \textsc{Acceptability} function which predicts the well-formedness of a shortening (\S\ref{s:extrinsic}), for use in application-specific compression systems.}
\end{itemize}

\section{Related work}\label{s:related_work}

Researchers have been studying extractive sentence compression for nearly two decades \cite{Knight2000StatisticsBasedS,clarke2008global,filippova2015sentence}. 
Recent approaches are often based on a large compression corpus,\footnote{\url{https://github.com/google-research-datasets/sentence-compression}} which was automatically constructed by using news headlines to identify ``gold standard'' shortenings \cite{filippova2013overcoming}. 
State-of-the-art models trained on this dataset \cite{filippova2015sentence,global_networks,Wang2017CanSH} can reproduce gold compressions (i.e.\ perfect token-for-token match) with accuracy higher than 30\%. 


However, because a sentence may be compressed in many ways (Table \ref{t:all_compressions}), this work introduces human acceptability judgements as a new and more flexible form of supervision for the sentence compression task.
Our approach is thus closely connected to research which seeks to model human judgements of the well-formedness of a sentence \cite{heilman2014predicting,shutzesprouse,lau2017grammaticality,Warstadt2018NeuralNA}. 
Unlike such studies, our work is strictly concerned with human perceptions of shortened sentences.\footnote{We compare our model to \citet{Warstadt2018NeuralNA} in \S\ref{s:intrinsic}.} 
Our work also solicits human judgements of shortenings from naturally-occurring news text, instead of sentences drawn from syntax textbooks \cite{sprouse2017design,Warstadt2018NeuralNA} or created via automatic translation \cite{lau2017grammaticality}. 

We note that our effort focuses strictly on anticipating the well-formedness of extractive compressions, rather than identifying compressions which contradict or distort the meaning of the original sentence. Identifying which compressions do not modify the meaning of source sentences is closely connected to the unsolved textual entailment problem, a recent area of focus in computational semantics \cite{snli_bowman,Pavlick2016SoCalledNA,linzencompression}.
In the future, we hope to apply this evolving research to the compression task. 
Some current compression methods use simple hand-written rules to guard against changes in meaning  \cite{clarke2008global}, or syntactic mistakes \cite{jing2000cut}. 

Finally, following much prior work, this study approaches sentence compression as a purely extractive task. Closely related work on abstractive compression \cite{cohn2008sentence,rush2015neural,mallinson18} and sentence simplification \cite{Zhu2010AMT,Xu2015ProblemsIC} seeks to shorten sentences via paraphrases or reordering of words.
Despite superficial similarity, extractive methods typically use different datasets, different evaluation metrics and different modeling techniques.

\section{Compression via subtree deletion}\label{s:framework}

Any sentence compression technique requires a framework for generating possible shortenings of a sentence, $s$. We generate compressions with a subtree deletion approach, based on prior work  \cite{Knight2000StatisticsBasedS,filippova2008dependency,filippova2013overcoming}. To generate a single compression (from among all possible compressions) we begin with a dependency parse of $s$.\footnote{We use Universal Dependency \cite{Nivre2016UniversalDV} trees (v1), parsed using CoreNLP \cite{corenlp}.} Then, at each timestep, we prune a single subtree from the parse. After $M$ subtrees are removed from the parse (one at a time, over $M$ timesteps) the remaining vertexes are linearized in their original order. 
Formally pruning a subtree refers to removing a vertex and all of its descendants from a dependency tree; pruning singleton subtrees (one vertex) is permitted.

We find that is possible to construct 88.2\% of the gold compressions in the training set of a standard corpus \cite{filippova2013overcoming} by only pruning subtrees. Therefore, we only examine prune-based compression in this work.\footnote{Extracting nested subclauses from source sentences is not possible with prune-only methods, because the root node of the compression must be the same as the root node of the original sentence. We plan to address this in future work.}

\section{Methodology and Dataset}\label{s:acceptability}

Our data collection methodology follows extensive research into human judgements of linguistic well-formedness \cite{shutzesprouse}. Such work has shown that non-expert participants offer consistent judgements of natural and unnatural sounding sentences with high test-retest reliability \cite{langsford2018quantifying}, across different data collection techniques \cite{bader_haussler_2010,sprouse2017design}. We apply this research to sentence compression, with confidence that our results reflect genuine human perceptions of shortened sentences because:
\begin{enumerate}
\itemsep0em 
\item{We carefully screen out many workers who offer judgements that violate well-known properties of English syntax, such as workers who approve deletion of objects from obligatory transitive verbs.}
\item{We observe that workers approve and disapprove of classes of sentences that English speakers would categorize as ``grammatical'' and ``ungrammatical,'' respectively. For instance, workers rarely endorse deletion of nominal subjects, and often endorse deletion of temporal modifiers.}
\item{Annotator agreement in our dataset is similar to agreement in prior, comprehensive studies of acceptability judgments (\S\ref{s:iaa}).}
\end{enumerate}

The appendix details our screening procedures, and discusses classes of accepted and rejected compressions in our dataset. 

\subsection{Measuring well-formedness}\label{s:acceptability_and_grammaticality}

Our study adopts a standard distinction between acceptability and grammaticality \cite{chomsky1965aspects,schutzebook}. 
Grammaticality is a binary and theoretical notion, used to characterize whether a sentence is or is not generable under a grammatical model. 
Acceptability is a measurement an individual's perception of the well-formedness of a sentence. 
Empirical studies have shown acceptability to be a gradient-based phenomenon, affected by a range of factors including plausibility, syntactic well-formedness, and frequency \cite{shutzesprouse}.
Based on this work, we expect that workers will have graded (not binary) perceptions of the well-formedness of compressions shown in our task. 

\begin{figure}[t]
\includegraphics[width=4.5cm]{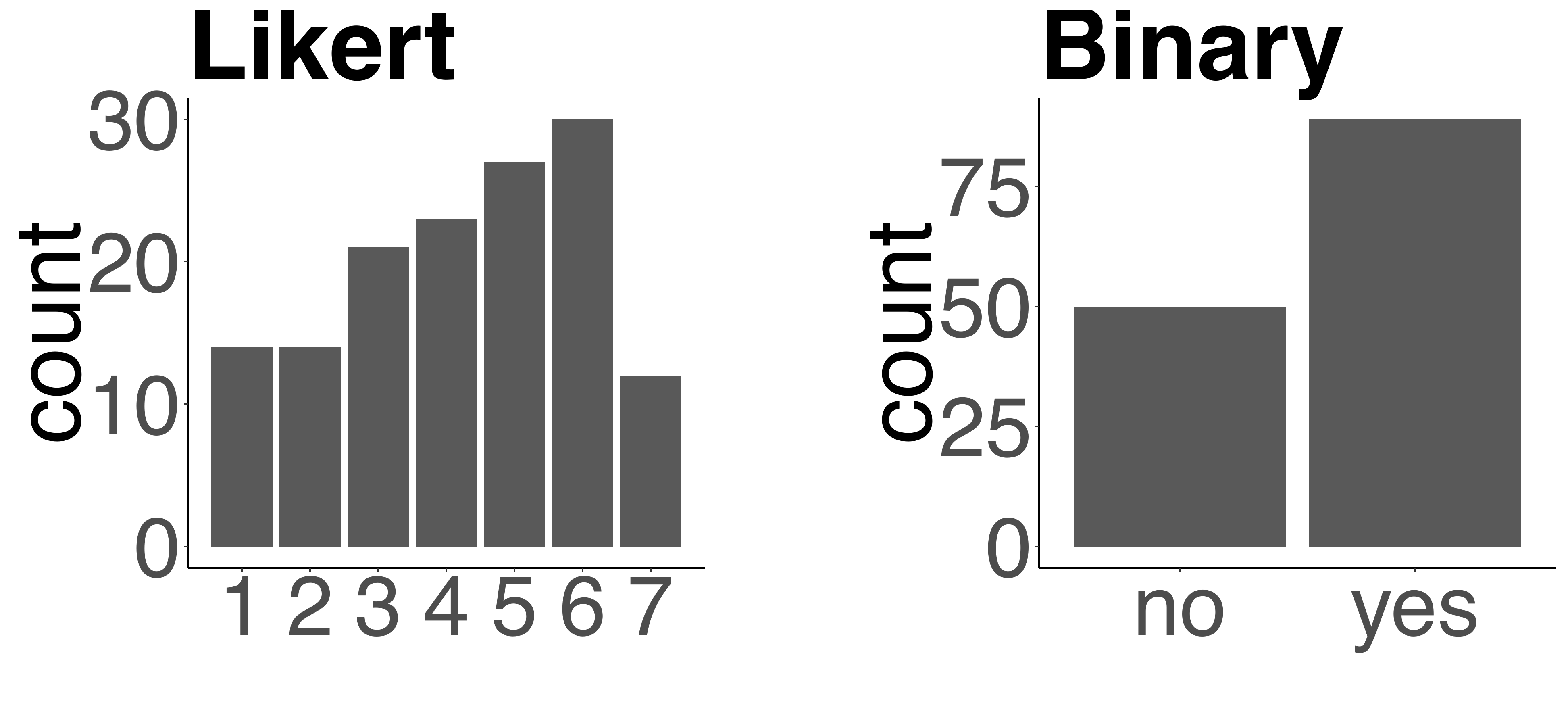}
\centering
\caption{Binary judgements and graded (Likert) judgements from \citet{sprouse2017design} for the slightly-awkward sentence, ``They suspected and we believed Peter would visit the hospital''. \citet{bader_haussler_2010} describe correlations between such measurement techniques.
}
\label{f:sprouse}
\end{figure}

Although acceptability is gradient-based, we nevertheless \textit{measure} worker perceptions by collecting binary judgements of well-formedness.
Earlier studies \cite{bader_haussler_2010,sprouse2017design,langsford2018quantifying} have shown that such binary measurements of acceptability correlate strongly with explicitly gradient collection techniques, such as Likert scales (Figure \ref{f:sprouse}).
We chose to collect binary judgements instead of graded judgments because (1)~binary judgments avoid ambiguity in how participants interpret a gradient scale~(2) binary judgements allowed us to write clear screener questions to block unreliable crowdworkers from the task and~(3) binary judgements allowed us to apply binary logistic modeling to directly predict observable worker behavior. 

\subsection{Data collection prompt}\label{s:prompt}

\begin{figure}[htb!]
\includegraphics[width=8cm]{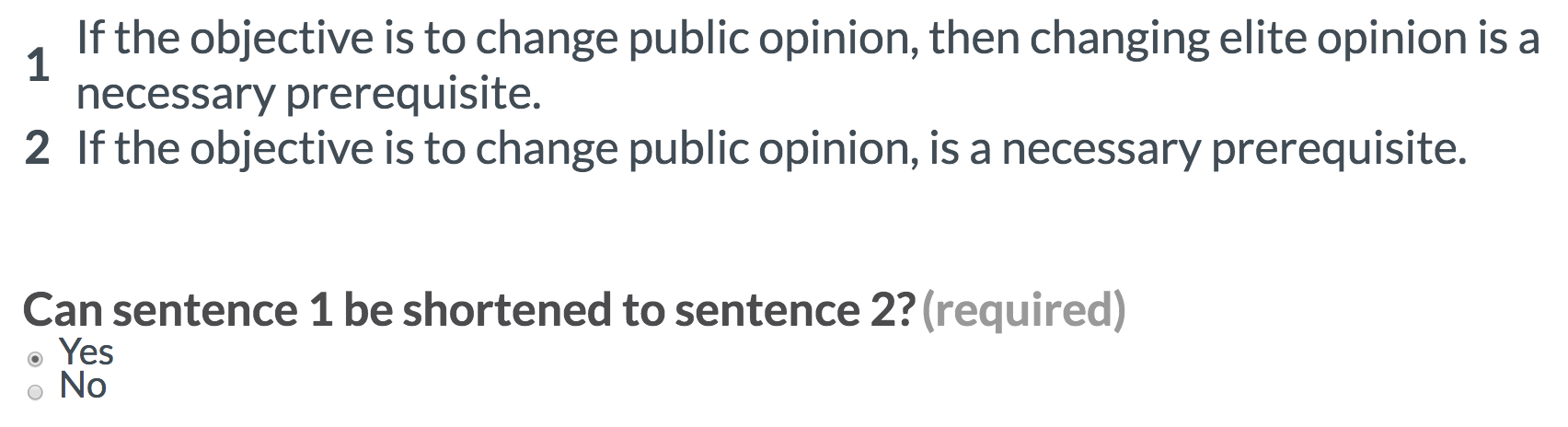}
\centering
\caption{Prompt to collect human judgements of acceptability for sentence compression. Workers are instructed to answer yes if the shorter sentence
``sounds good'' or ``sounds like something a person would say.'' }
\label{f:task}
\end{figure}

We show crowdworkers on Figure Eight\footnote{\url{https://www.figure-eight.com/}.} a naturally-occurring sentence, along with a proposed compression of that same sentence, generated by executing a \textit{single} prune operation on the sentence's dependency tree.
We then ask a binary question: can the longer sentence be shortened to the shorter sentence?
Our prompt is shown in Figure \ref{f:task}. 
We instruct workers to say yes if the shorter sentence ``sounds good'' or ``sounds like something a person would say,'' following verbiage for the acceptability task \cite{shutzesprouse}. 

Because we designed our task to follow typical acceptability prompts, we expect that workers completing the task evaluated the wellformedness of each compressed sentence, and then answered \textit{yes} and if they deemed it acceptable. 

We instruct workers to say yes if a compression sounds good, even if it changes the meaning of a sentence. 
While practitioners will need to identify shortenings which are both syntactically well-formed \textit{and} which do not modify the meaning of a sentence, this work focuses strictly on identifying well-formed compressions.
In the future, we plan to apply active research in semantics (\S\ref{s:related_work}) to identify disqualifying changes in meaning.

\subsection{Dataset details}\label{s:properties}


We generate 10,128 sentence--compression pairs from a freely-distributable corpus of web news \cite{voxcorpus}. Each source sentence $s$ is chosen at random, and each compression $c$ is produced by a single, randomly-chosen prune operation on $s$. Our data thus reflects the natural distribution of dependency types in the corpus.

\begin{table}[htb!]
\begin{tabular}{@{}ll@{}}
\toprule
$N$ judgements (train) & 6010 {\small (4522 sents.)}  \\
$N$ judgements (test) & 640 {\small (486 sents.)} \\
class balance & 64.2\%/35.8\%  {\small (no/yes)}    \\
overall compression rate          & $\mu$=0.867 {\small $\sigma$=0.174}  \\ \bottomrule
\end{tabular}
\caption{Corpus statistics.}
\label{t:corpus_stats}
\end{table}

We present each $(s,c)$ pair to 3 or more workers,\footnote{FigureEight will sometimes solicit additional judgements automatically.} then conservatively exclude many judgements from workers who are revealed to be inattentive or careless (see appendix), in order to be certain that worker disagreements in our dataset reflect genuine perceptions of well-formedness. We then divide filtered data into a training and test set by $(s,c)$ pair, so that our model does not use a train-time judgement about $(s,c)$ from worker $k$ to predict a test-time judgement about $(s,c)$ from worker $k^{\prime}$. Table \ref{t:corpus_stats} presents dataset statistics.\footnote{Note that we use a character-based \citet{filippova2015sentence} rather than token-based \cite{napoles2011evaluating} definition of compression rate.} Our organization does not require institutional approval for crowdsourcing. 

\subsection{Inter-annotator agreement}\label{s:iaa}

There are at least two sources of inter-annotator disagreement which could arise in our data. First, in cases when a compression is neither entirely garbled nor perfectly well formed, previous empirical studies \cite{sprouse2017design,langsford2018quantifying} suggest that annotators will likely disagree. (See Figure \ref{f:sprouse} and \S\ref{s:acceptability_and_grammaticality}). Second, we suspect that different individuals set different thresholds on how acceptable a sentence must be before they give a ``yes'' response: a compression that one person rates as acceptable might be rated by the next as unacceptable, even if they have the same impression of the compression's acceptability. Such between-annotator variability might represent a form of response bias, which is common in psychological experiments \cite{macmillan1990response,macmillan2005detection}. We attempt to control for such bias by including a worker ID feature in our model (\S\ref{s:features}).

To evaluate the extent of such disagreement, and to compare with other work, we measure inter-annotator agreement using Fleiss' kappa \cite{fleiss1971measuring}, computing a $\kappa=.294$ on the entire filtered dataset.\footnote{See appendix: details of Fleiss $\kappa$ for crowdsourcing.} 
We observe a similar rate of agreement (${\kappa=.323}$) in a comprehensive study of acceptability judgements \cite{sprouse2017design}.\footnote{We compute this number using publicly released data from the YN study.} Our $\kappa$ is lower than typical in standard annotation paradigms in NLP, which often attempt to assign instances to hard classes, rather than measure graded phenomena.





\section{Intrinsic Task: Modeling single operations}\label{s:intrinsic}

We create a model to predict if a given worker will judge that a given single-operation compression is acceptable. We say that $Y=1$ if worker $k$ answers that $s$ can be shortened to $c$. We then model $p(Y = 1 |\bm{W},  \bm{x})$ using binary logistic regression, where $\bm{x} = \phi(s, c, k)$ is a feature vector reflecting the nature of the edit which produces $c$ from $s$, as observed by worker $k$.

\subsection{Model Features}\label{s:features}

The major features in our model are: $(i)$ language model features, $(ii)$ dependency type features, $(iii)$ worker ID features, $(iv)$ features reflecting properties of the edit from $s$ to $c$ and $(v)$ interaction features. We discuss each below.



 \textbf{Language model features}. Our model builds upon earlier work examining the relationship between language modeling and acceptability judgements. In a prior study, \newcite{lau2017grammaticality} define several functions which normalize predictions from a language model by token length and by word choice; then test which functions best align with human acceptability judgements.
 We use their Norm LP function in our model, defined as:

\begin{equation}
\text{Norm LP} (\xi) \triangleq -\frac{\text{log } p_m(\xi)}{\text{log } p_u(\xi)}
\end{equation}

where $\xi$ is a sentence, $p_m(\xi)$ is the probability of $\xi$ given by a language model and $p_u(\xi)$ is the unigram probability of the words in $\xi$.

We use Norm LP  as a part of two features in our approach. One real-valued feature records the probability of a compression computed by Norm LP($c$). Another binary feature computes Norm LP($s$) - Norm LP($c$) $>$ 0. The test set performance of these language model (LM) features is shown in Table \ref{t:bigtable}. The appendix further describes our implementation of Norm LP. 

\textbf{Dependency type features.} We use the dependency type governing the subtree pruned from $s$ to predict the acceptability of $c$. This is because workers are more likely to endorse deletion of certain dependency types. For instance, workers will often endorse deletion of temporal modifiers, and often reject deletion of nominal subjects.

\textbf{Worker ID features}. We also include a feature indicating which of the workers in our study submitted a particular judgement for a given $(s,c)$ pair.
We include this feature because we observe that different workers have greater or lesser tolerance for more and less awkward compressions.\footnote{More formally, we define a given worker's deletion endorsement rate as the number of times a worker answers yes, divided by their total judgements. 
We observe a roughly normal distribution ${\tiny (\mu=.402 \text{ }, \sigma=.216)}$ of worker deletion endorsement rates across the dataset.}

Including the worker ID feature allows our model to partially account for an individual worker's judgement based on their overall endorsement threshold, and partially account for a worker's judgement based on the linguistic properties of the edit. 
The feature thus controls for variability in each worker's baseline propensity to answer yes. 
Because real applications will not have access to worker-specific information, we do not use the worker ID feature in evaluating our model and dataset for use in practical compression systems (\S \ref{s:extrinsic}). All workers in the test set submit judgements in the training set.  

\begin{table*}[]
\centering
\begin{tabular}{lccl}
 & \multicolumn{2}{c}{Hard classification ($t$=0.5) } & \multicolumn{1}{c}{Ranking} \\
  \cmidrule(r){2-3}
  \cmidrule(r){4-4}
Model  & Accuracy  & Fleiss $\kappa$ & ROC AUC $(p)$ \\ \cline{1-4}
CoLA & 0.622 &  -0.210 & 0.590 {\footnotesize$(<.001)$} 							\\ \cline{1-4}
language model (LM)&0.623&-0.232&0.583 {\footnotesize$(<.001)$}\\
+ dependencies&0.664&0.124&0.646 {\footnotesize$(<.001)$}\\
+ worker ID&0.695&0.232&0.746 {\footnotesize$(<.001)$}\\ \hline
full $\triangleq p(Y = 1 |\bm{W}, \bm{x})$& \textbf{0.742}&\textbf{0.400}&\textbf{0.807}  \\
-  dependencies&0.731&0.368&0.797 {\footnotesize$(0.073)$}\\ 
-  worker ID&0.667&0.170&0.691 {\footnotesize$(<.001)$}\\ \hline
worker -- worker agreement & 0.636$^{*}$  & 0.270 &--- \\
\citet{sprouse2017design} & ---  &  0.323 &--- \\
\end{tabular}
\caption{Test set accuracy, Fleiss' $\kappa$ and ROC AUC scores for six models, trained on the single-prune dataset (\S\ref{s:acceptability}), as well as scores for a model trained on the CoLA dataset \protect\cite{Warstadt2018NeuralNA}. The simplest model uses only language modeling (LM) features. We add dependency type (+  dependencies) and worker ID (+ worker IDs) information to this simple model. We also remove dependency information (-  dependencies) and worker information  (-  worker ID) from the full model. The full model achieves the highest test set AUC; $p$ values beside each smaller AUC score show the probability that the full model's gains over the smaller AUC score occurs by chance. We also compute $\kappa$ for each model by calculating the observed and pairwise agreement rates \protect\cite{fleiss1971measuring} for judgements submitted by the crowdworker and ``judgements'' submitted by the model. Models which can account for worker effects achieve higher accuracies than the observed agreement rate among workers ($0.636^{*}$), leading to higher $\kappa$ than for worker--worker pairs.}
\label{t:bigtable}
\end{table*} 

\textbf{Edit property features.}  We also include several features which register properties of an edit, such as features which indicate if an operation removes tokens from the start of the sentence, removes tokens from the end of a sentence, or removes tokens which follow a punctuation mark. We include a feature that indicates if a given operation breaks a collocation (e.g.\ ``He hit a \sout{home} run''). The appendix details our collocation-detection technique. 

\textbf{Interaction features.} Finally, we include seventeen interaction features formed by crossing exit property features with particular dependency types. For instance, we include a feature which records if a prune of a \rdep{conj} removes a token following a punctuation mark.




\subsection{Evaluation}\label{s:eval}

We compare our model of individual worker judgements to simpler approaches which use fewer features (Table \ref{t:bigtable}), including an approach which uses only language model information \cite{lau2017grammaticality} to predict acceptability. We compute the test set accuracy of each approach in predicting binary judgements from individual workers, which allows for comparison with agreement rates between workers.  However, because acceptability is a gradient-based phenomenon (\S\ref{s:acceptability}), we also evaluate without an explicit decision threshold via the area under the receiver operating characteristic curve (ROC AUC),
which measures the extent to which an approach ranks good deletions over bad deletions. ROC AUC thus measures how well predicted probabilities of binary positive judgements correlate with human perceptions of well-formedness. Other work which solicits gradient-based judgements instead of binary judgements evaluates with Pearson correlation \cite{lau2017grammaticality};
ROC AUC is a close variant of the Kendall ranking correlation \citep{Newson2002Kendall}. Our full model also achieves a higher AUC than approaches that remove features from the model.  We use bootstrap sampling \cite{D12-1091} to test the significance of AUC gains (Table \ref{t:bigtable}); $p$ values reflect the probability that the difference in AUC between the full model and the simpler model occurs by chance. 

%
%

	
The probability that two workers, chosen at random, will agree that a given $s$ in the test set may be shortened to a given $c$ in the test set is 63.6\%. We hypothesize that the full model's accuracy of 74.2\% is higher than the observed agreement rate between workers because the full model is better able to predict if worker $k$ will endorse an individual deletion. 

We also compare our full model to a baseline neural acceptability predictor trained on CoLA \cite{Warstadt2018NeuralNA}: a corpus of grammatical and ungrammatical sentences drawn from syntax textbooks. Using a pretrained model, we predict the probability that each source sentence and each compression is well formed, denoted CoLA$(s)$ and CoLA$(c)$. We use these predictions to define four features: CoLA$(c)$, log CoLA$(c)$, CoLA$(s)$ - CoLA$(c)$, and log CoLA$(s)$ - log CoLA$(c)$. We show the performance of this model in Table \ref{t:bigtable}. CoLA's performance for extractive compression results warrants future examination: large corpora designed for neural methods sometimes contain limitations which are not initially understood \cite{chen_CNN,liang_squad,annotation_artifacts_snli}. 


\section{Extrinsic Task: Modeling multi-operation compression}\label{s:extrinsic}

In this work, we argue that a single sentence may be compressed in many ways; a ``gold standard'' compression is just one of the exponential possible shortenings of a sentence. Instead of mimicking gold standard training data, we argue that compression systems should support application-specific  \textsc{Brevity} constraints and \textsc{Importance} criteria: stock traders and stylists will have different definitions of ``important'' content in fashion blogs, compressions for a desktop will be longer than compressions for a phone.

Nevertheless, in many application settings, compression systems will try to show users well-formed shortenings. Thus, in the remainder of this study, we examine how our transition-based sentence compressor (\S\ref{s:framework}), supervised with human acceptability judgements (\S\ref{s:intrinsic}), may be be used to provide $\textsc{Acceptability}$ scores which align with human perceptions of well-formedness.\footnote{Early approaches to sentence compression used language models \cite{clarke2008global} or corpus statistics \cite{filippova2008dependency} to generate ``readability'' scores. Newer neural approaches \cite{filippova2015sentence} rely on implicit definitions of well-formedness encoded in training data.}  Such scores could be used as a component of many different practical sentence compression systems, including a method described in \S{\ref{s:single}}. 

\subsection{\textsc{Acceptability} scores}

We consider any function which maps a compression to some real number reflecting its well-formedness to be an $\textsc{Acceptability}$ score. In \S\ref{s:intrinsic}, we present a model, $p(Y = 1 |\bm{W},  \bm{x})$, which attempts to predict which operations on well-formed sentences return reasonable compressions. If we execute a chain of $M$ such operations, and assume that each operation's effect on acceptability is independent, we can model the probability that $M$ prune operations will result in an acceptable compression with $\prod_i^M p(Y = 1 |\bm{W},  \bm{x_i})$, which is equal to the chance that a person will endorse each of the $M$ deletions. We test this model with a function that expresses the probability that all operations are acceptable: 

\begin{equation}
\textsc{A} (c) \triangleq \sum_{i=1}^{M} \text{log~~}  p(Y = 1 |\bm{W},  \bm{x_i})
\end{equation}\label{eq:main}

\noindent where each $\bm{x_i}$ are features reflecting the nature of the prune operations which shortens $c_i$ to $c_{i + 1}$ in the chain of operations, and where ${p(Y = 1 |\bm{W},  \bm{x_i})}$ is the predicted probability of deletion endorsement under our full model. (Because no worker observes the deletion, we do not use the worker ID feature in predicting deletion endorsement.)

The sum of log probabilities in $\textsc{A}(c)$ reflects the fact that any operation on a well-formed sentence carries inherent risk: modifying a sentence's dependency tree may result in a compression which is not acceptable. The more operations executed the greater the chance of generating a garbled compression. We use this intuition to define a simpler alternative, $\textsc{A}_{M}(c) \triangleq -M$, where $M$ is the number of prune operations used to create the compression. We also examine a function $\textsc{A}_{\text{min}}(c) \triangleq \text{min} \{ \text{log }  p(Y = 1 | \bm{W},  \bm{x_i}) \mid i \in 1 .. M \}$, which represents our observation that a single operation with a low chance of endorsement will often create a garbled compression. We compare these functions to $\textsc{A}_{LM}(c)$, which is equal to the probability of the compression $c$ under a language model, normalized by sentence length. (We use the Norm LP formula defined in \S\ref{s:intrinsic}. Language model predictions have been shown to correlate with the well-formedness of a sentence \cite{lau2017grammaticality,kannConl}.) Finally, we test a function $\textsc{A}_{CoLA}(c)$, which is equal to the predicted probability of well-formedness of the compression from a pretrained acceptability predictor (\S\ref{s:intrinsic}).

\subsection{Experiment}\label{s:multi_op_experiment}

To evaluate each \textsc{Acceptability} function, we collect a small dataset of multi-prune compressions.\footnote{Rather than single-prune compressions (\S\ref{s:acceptability}).} We draw 1000 initial candidate sentences from a standard compression corpus \cite{filippova2013overcoming}, and then remove sentences which are shorter than 100 characters to create a sample of 958 sentences. We then compress each of the sentences in the sample by executing prune operations in succession until the character length of the remaining sentence is less than $B$, a randomly sampled integer between 50 and 100. This creates an evaluation dataset with a (roughly) uniform distribution, by character length.

To generate each compression, we use a sampling method which allows us to explore a wide range of well-formed and garbled shortenings, without generating too many obviously terrible compressions.\footnote{Very many exponential possible compressions of a single sentence will be garbled or non-sensical. Pruning even a single subtree at random from an acceptable sentence (\S\ref{s:intrinsic}) destroys acceptability more than 60\% of the time.} 
Concretely, we sample each prune operation in each chain in proportion to $p(Y=1)$, a model's prediction (\S\ref{s:intrinsic}) that the edit will be judged acceptable. This means that we delete vertex $v$ and its descendants with probability $\frac{1}{Z}p(y=1 | W, s, c_v)$, where $c_v$ is the compression formed by pruning the subtree rooted at $v$ and $Z=\sum_{v \in \mathcal{V}} p(y=1 | W, s, c_v)$ is the sum of the endorsement probability of all possible compressions.\footnote{In the behavioral sciences, this method of choosing actions is sometimes called \textit{probability matching} \cite{vulkan2000economist}.} 

We show each sentence in the evaluation dataset to 3 annotators, using our acceptability prompt (Figure \ref{f:task}). This creates final evaluation set consisting of 2,388 judgements of 940 multi-prune compressions, after we implement the judgement filtering process described in the appendix. We compute $\kappa$=0.099 for the evaluation dataset.


For all defined \textsc{Acceptability} functions, we measure AUC against binary worker deletion endorsements (yes or no judgements) in the evaluation dataset to determine the quality of the ranking produced by each \textsc{Acceptability} function (\S\ref{s:eval}). The function $\textsc{A}(c)$, which integrates information from a language model, as well as information about the grammatical details of the process which creates $c$ from $s$, achieves the highest AUC on the evaluation set, best correlating with human judgements of well-formedness. 

\begin{table}[]
\centering
\begin{tabular}{l l c}
\cline{1-3}
 Function & Description &   ROC AUC  \\  \midrule
 $\textsc{A}_{CoLA}(c)$ &  {\footnotesize CoLA pretrained} & .510  \\
 $\textsc{A}_{LM}(c)$ &  {\footnotesize Language model} & .557  \\
 $\textsc{A}_{M}(c)$ & {\footnotesize Number of operations}  $\times$ -1& .580  \\
 $\textsc{A}_{\text{min}}(c)$&  {\footnotesize Least acceptable operation}   & .581 \\
 $\textsc{A} (c)$ & {\footnotesize All operations acceptable} & .591 
\end{tabular}
\caption{ROC AUC for several \textsc{Acceptability} functions for the multi-operation compression task. The $\textsc{A} (c)$ model achieves a gain of .034 in AUC over the $\textsc{A}_{LM}(c)$ model {\scriptsize $ (p=0.005)$. }}
\end{table}



\subsection{One sentence, many compressions}\label{s:single}

This work argues that there is no single best way to compress a sentence. We demonstrate this idea by examining some of the exponential possible compressions of the sentence shown below. (This same sentence is also shown in Table \ref{t:all_compressions}.)

\begin{table}[htb!]
\centering
\begin{tabular}{p{7cm}}
$s$ = Pakistan launched a search for its missing ambassador to Afghanistan on Tuesday, a day after he disappeared in a Taliban area. \\ 
$c_{g}$ = Pakistan launched a search for its missing ambassador \\ 
$c_1$ = Pakistan launched a search for its missing ambassador to Afghanistan on Tuesday \\ 
$c_2$ = Pakistan launched search Tuesday \\
\end{tabular}
\caption{A sentence $s$ and a ``gold'' compression ($c_g$) from a standard corpus \cite{filippova2013overcoming}, along with two alternate compressions ($c_1$ and $c_2$).}
\label{t:single_sentence}
\end{table}

We generate an initial list of 1000 possible compressions of this 126-character sentence via the procedure defined in \S\ref{s:multi_op_experiment}, and we score the \textsc{Acceptability} of each compression by $\textsc{A}(c)$. In this instance, we define \textsc{Brevity}$(c)$ to be the maximum character length of a compression and \textsc{Importance}$(c)$ to be a binary function returning 1 only if the compression includes the query term, ``Afghanistan''. (Practical compression systems would also need to check for changes in meaning resulting from deletion (\S\ref{s:related_work}), but we leave this step for future work.)

\begin{figure}[htb!]
\includegraphics[width=5cm]{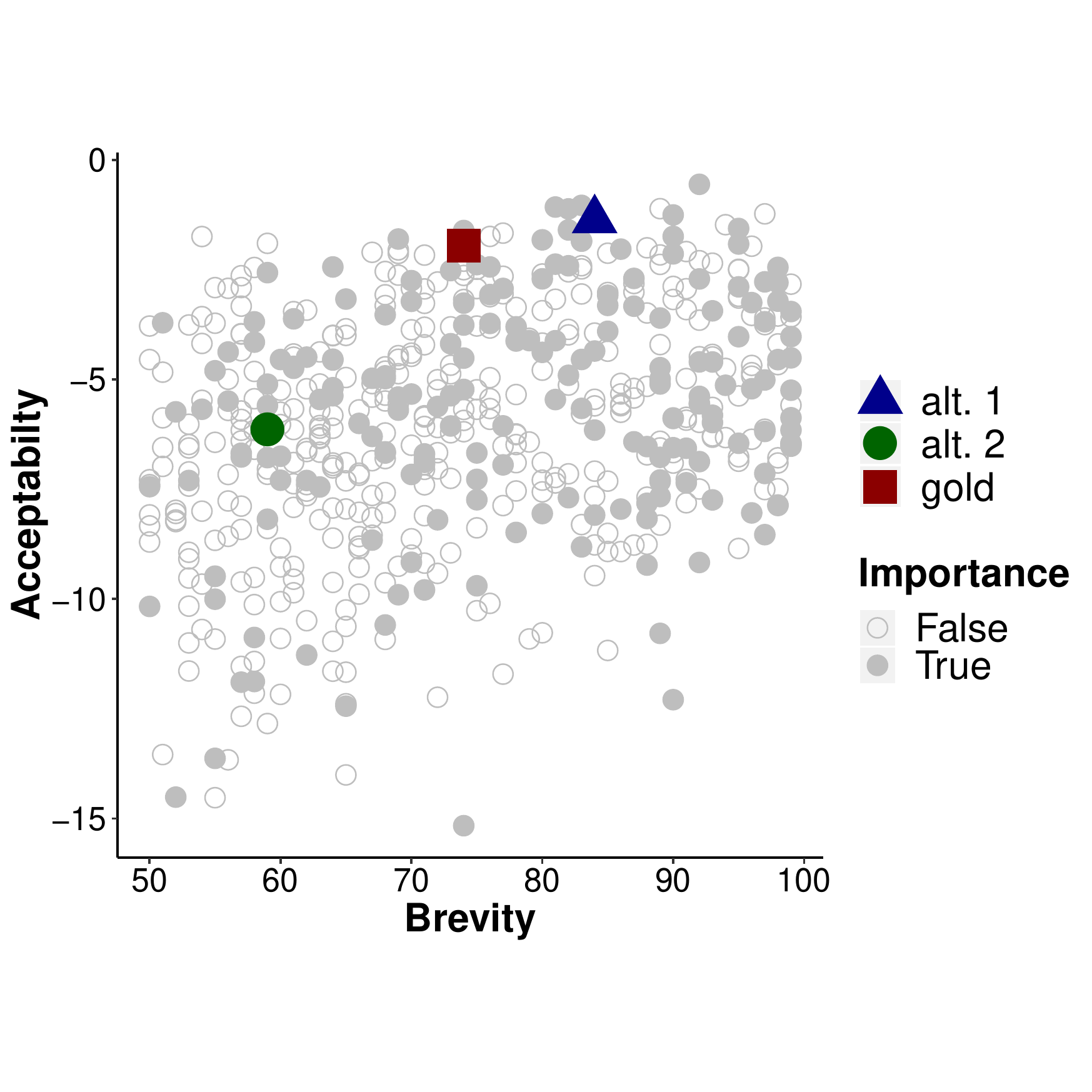}
\centering
\vspace{-.5cm}
\caption{554 possible compressions of a single sentence (Table \ref{t:single_sentence}), displayed by  $A(c)$ score, \textsc{Brevity} constraint and \textsc{Importance} criterion. The ``gold standard'' compression, $c_g$, is shown with a large red square, along with alternate $c_1$ (large triangle), and alternate $c_2$ (large circle).}
\label{f:single_sentence}
\end{figure}

Following deduplication steps described in the appendix, we generate a final list of 554 different possible compressions of the sentence. We plot each of the 554 compressions in Figure \ref{f:single_sentence}, which shows many possible shortenings with high $A(c)$ scores. The ``gold standard'' compression is just one arbitrary shortening of a sentence.



\section{Conclusion and future work}\label{s:conclusion}

Our effort suggets areas for future work. To begin, our study is strictly concerned with grammatical and not semantic acceptability. In the future, we hope to apply active work from semantics (\S{\ref{s:related_work}}) to identify meaning-changing compressions. We also plan to add support for additional operations such as paraphrasing or forming compressions from nested subclauses. In addition, we plan to develop improved models of multi-prune compression, and to apply our \textsc{Acceptability} scores in loss functions for neural compression techniques. 

\section{Appendix}\label{s:appendix}



\subsection{Crowdsourcing details}

We paid workers 5 cents per judgement and only opened our task to US-based workers with a level-2 designation on Figure Eight. 

Following standard practice on the crowdsourcing platform, we used test questions to screen out careless workers \cite{snow2008cheap}. All workers began our job with a \textit{quiz mode} of 10 screener questions, and then saw one screener question in every subsequent page of 10 judgements. Workers who failed more than 80\% of test questions were screened out from the task. Screener questions were indistinguishable from our regular collection prompt. 

We wrote test screener questions based on established understanding of English syntax, to avoid biasing results with our own subjective judgements of acceptability.
For example, linguists have extensively examined which English verbs require objects and which verbs do not require objects via corpus-based, elicitation-based and eye-tracking methods \cite{gahl2004verb,eye}. 
We used this work to write screener questions which check that workers answer no for operations that prune direct objects of obligatory transitive verbs. Similarly, we wrote screener questions which check that workers answer no to deletions which split a verb and a known obligatory particle in a multiword expression \cite{Baldwin2010MultiwordE}, or remove determiners before singular count nouns \cite[p.\ 354]{huddleston2002the}. 
For the multi-prune dataset (\S\ref{s:multi_op_experiment}), we added test questions which confirmed that workers approved of well-formed, gold standard compressions from a standard corpus \cite{filippova2013overcoming}.

We also include screener questions which check if a worker is paying attention, along with several poll questions which ask workers if they grew up speaking English.\footnote{Workers are instructed there is no right answer to questions about language background, so there is no incentive to answer dishonestly.}
We ignore judgements from known non-native speakers and known inattentive workers in downstream analysis.\footnote{We also exclude 1663 suspected fraudulent judgements from 17 IP address associated with multiple worker IDs.}
We defined rules for filtering the dataset before examining the test set to ensure that filtering decisions did not influence test-set results.
We release screener questions and task instructions along with crowdsourced data for this work.

\subsection{Per-dependency deletion endorsements}

Breaking out worker responses by dependency type provides additional validation for our data collection approach. 
We observe that workers are unlikely to endorse deletion of dependency types which create compressions that English speakers would deem ``ungrammatical,'' and likely to endorse deletions which speakers would deem ``grammatical.''

For example, in UD, the \rdep{mwe} relation is most commonly used to link two or more function words that obligatorily occur together (e.g.\ \textit{because of, due to, as well as}). Since deleting a \rdep{mwe} amounts to suppressing a critical closed-class item, it is not surprising that, overall, workers only assented to deleting   \rdep{mwe} in 9.5\% of cases. Similarly, low deletion endorsement for the \rdep{cop} relation ({\small 15.6\%}) is consistent with the grammatical rules of mainstream varieties of American English, which generally require an overt copula in copular constructions.\footnote{Not all dialects require overt copulas \cite{green2002african}.} 

On the other hand, we found that optional \cite{larson1985syntax} pre-conjunction operators like \textit{both} or \textit{either} were almost always considered removable ({\small 80.0\%} deletion endorsement). Workers also endorsed the deletion of temporal adverbs such as \textit{tomorrow} or \textit{the day after next} 78.9\% of the time, which is sensible as temporal adverbs are typically considered adjuncts. 

Since these response patterns generally align with well-established grammatical generalizations \cite{huddleston2002the}, they serve to validate our data collection approach. 

\subsection{Experimental details}

We report additional details regarding several of the experiments in the paper, presented in the order in which experiments appear.

\textbf{Fleiss $\kappa$.} Fleiss' original metric \cite{fleiss1971measuring} assumes that each judged item will be judged by exactly the same number of raters. 
However, our data filtering procedures create a dataset with a variable number of raters per sentence-compression pair. 
(This is common in crowdsourcing.) 
We thus calculate the observed agreement rate for an individual item ($P_i$, in Fleiss' notation) by computing the pairwise agreement rate from among all raters for that item. 
We ignore cases where only one rater judged a given $(s,c)$ pair, which occurs for 73.2\% of pairs. 

\textbf{Tuning and implementation.} We implement our model $\triangleq p(Y = 1 |\bm{W}, \bm{x})$ with scikit-learn \cite{Pedregosa:2011:SML:1953048.2078195} using L2 regularization. We tune the inverse regularization constant to $c=0.1$ to optimize ROC AUC in 5-fold cross validation over the training set, after testing $c \in\{10^j \mid j \in \{-3,-2,-1,0,1,2 \} \}$. We do not include a bias term. All other settings are set to default values.

\textbf{NormLP.} Following \citet{lau2017grammaticality}, in this work, we use the Norm LP function to normalize output from a language model to predict grammaticality. 
Our Norm LP function uses predictions from a 3-gram language model trained on English Gigaword \cite{gigaword5} and implemented with KenLM \cite{kenlm}. \newcite{lau2017grammaticality} report identical performance for the Norm LP function using 3-gram and 4-gram models.

\citet{lau2017grammaticality} found that another function, SLOR$(\xi)$, performed as well as Norm LP in predicting human judgements. 
We found that Norm LP achieved higher AUC than SLOR in 5-fold cross-validation experiments with the training set.\footnote{\citet{kannConl} also examine SLOR for automatic fluency evaluation.}

\textbf{Collocations.} Our model includes a binary feature, indicating if an edit breaks a collocation. We identify collocations by computing the offsets (signed token distances) between words \cite[ch. 5.2]{manning1999foundations} in English Gigaword \cite{gigaword5}. If the variance in token distance between two words is less than 2 and the mean token distance between the words is less than 1.5 we deem the words a collocation. We identify 647 total edits (across train and test sets) which break a collocation; only 11 of such edits are for \rdep{mwe} relations. Examples include: ``forget \sout{about it}''', ``kind \sout{of}'' and ``as \sout{well}''.

\textbf{CoLA.} All reported results for the CoLA model use the Real/Fake + LM (ELMo) baseline from \citet{Warstadt2018NeuralNA}.\footnote{\url{https://github.com/nyu-mll/CoLA-baselines}} Across our entire dataset, the mean predicted acceptability of source sentences from the CoLA model is 0.867 ($\sigma$=0.264) and the predicted acceptability of compressions is 0.740 ($\sigma$ = 0.363).  We hypothesize that compression scores have a greater variance and a lower mean because only some compressions are well-formed.


\textbf{Deduplication of possible compressions.} In this work, we describe a method for generating multiple compressions from a single sentence. In our generation procedure, it is possible to randomly select the exact same sequence of operations multiple times. During these experiments, we remove any such duplicates from the initial list. 

Additionally, in our compression framework, the sequence of operations which produces a given shortening is not unique.\footnote{For instance, it is possible to prune a leaf vertex with one operation, and then prune its parent vertex with a second operation; or just remove both vertexes at once via a single prune of the parent. (Each sequence returns the same compression.)} In cases where different sequences of operations return the same compression, we select the compression with the highest $\textsc{A}(c)$ score, which represents the best available path to the shortening.

\bibliography{abe}
\bibliographystyle{acl_natbib}

\end{document}